\documentclass{article}

\usepackage{microtype}
\usepackage{graphicx}
\usepackage{subcaption}
\usepackage{booktabs} 
\usepackage[table]{xcolor}  

\usepackage[hyphens]{url}
\usepackage{hyperref}

\usepackage[accepted]{icml2021}

\icmltitlerunning{Orthonormality and Sparsity for Molecular Graph Representations}

\begin{document}

\twocolumn[
\icmltitle{Improving Molecular Graph Neural Network Explainability \\
           with Orthonormalization and Induced Sparsity}




\begin{icmlauthorlist}
\icmlauthor{Ryan Henderson}{bayer}
\icmlauthor{Djork-Arn\'{e} Clevert}{bayer}
\icmlauthor{Floriane Montanari}{bayer}
\end{icmlauthorlist}

\icmlaffiliation{bayer}{Digital Technologies, Bayer AG, Berlin, Germany}

\icmlcorrespondingauthor{Ryan Henderson}{ryan.henderson@bayer.com}
\icmlcorrespondingauthor{Floriane Montanari}{floriane.montanari@bayer.com}

\icmlkeywords{Machine Learning, ICML}

\vskip 0.3in
]



\printAffiliationsAndNotice{}  

\begin{abstract}
    Rationalizing which parts of a molecule drive the predictions of a molecular graph convolutional neural network (GCNN) can be difficult.
    To help, we propose two simple regularization techniques to apply during the training of GCNNs: Batch Representation Orthonormalization (BRO) and Gini regularization.
    BRO, inspired by molecular orbital theory, encourages graph convolution operations to generate orthonormal node embeddings.
    Gini regularization is applied to the weights of the output layer and constrains the number of dimensions the model can use to make predictions.
    We show that Gini and BRO regularization can improve the accuracy of state-of-the-art GCNN attribution methods on artificial benchmark datasets.
  In a real-world setting, we demonstrate that medicinal chemists significantly prefer explanations extracted from regularized models.
    While we only study these regularizers in the context of GCNNs, both can be applied to other types of neural networks.

\end{abstract}

\section{Introduction}\label{sec:introduction}

Graph convolutional neural networks (GCNNs) have shown particular promise in predicting properties of small molecules. 
These properties can be biological activity against protein targets~\cite{yang_comprehensive_2020, sakai_prediction_2021, nguyen_meta-learning_2020}, more general pharmacokinetics and physicochemical properties~\cite{montanari_modeling_2019, peng_enhanced_2020, feinberg_improvement_2020} or toxicity~\cite{ma_deep_2020}.
Using accurate \textit{in silico} predictions of such properties to prioritize the synthesis of compounds can lead to tremendous time and cost savings during drug discovery projects.

However, the opaque nature of artificial neural networks has long been a stumbling block for wider adoption.
In particular, the difficulty in extracting a straightforward rationalization for a molecular prediction in terms of atomic or fragment contributions limits the adoption of machine learning models among medicinal chemists.

What makes a good rationalization?
In the case of explaining molecular properties, the answer is often site attribution.
A chemist, when presented with an image of a molecule, may infer properties like solubility and melting point from a few specific atoms and fragments rather than the molecule as a whole.
Explaining the predictions in terms of atomic contributions helps build trust in deep learning for molecular properties.

\section{Our Contribution}\label{sec:our-contribution}

\begin{figure*}[t]
    \vskip 0.2in
    \begin{center}
        \centerline{\includegraphics[trim=0 100 0 100, clip, width=\textwidth]{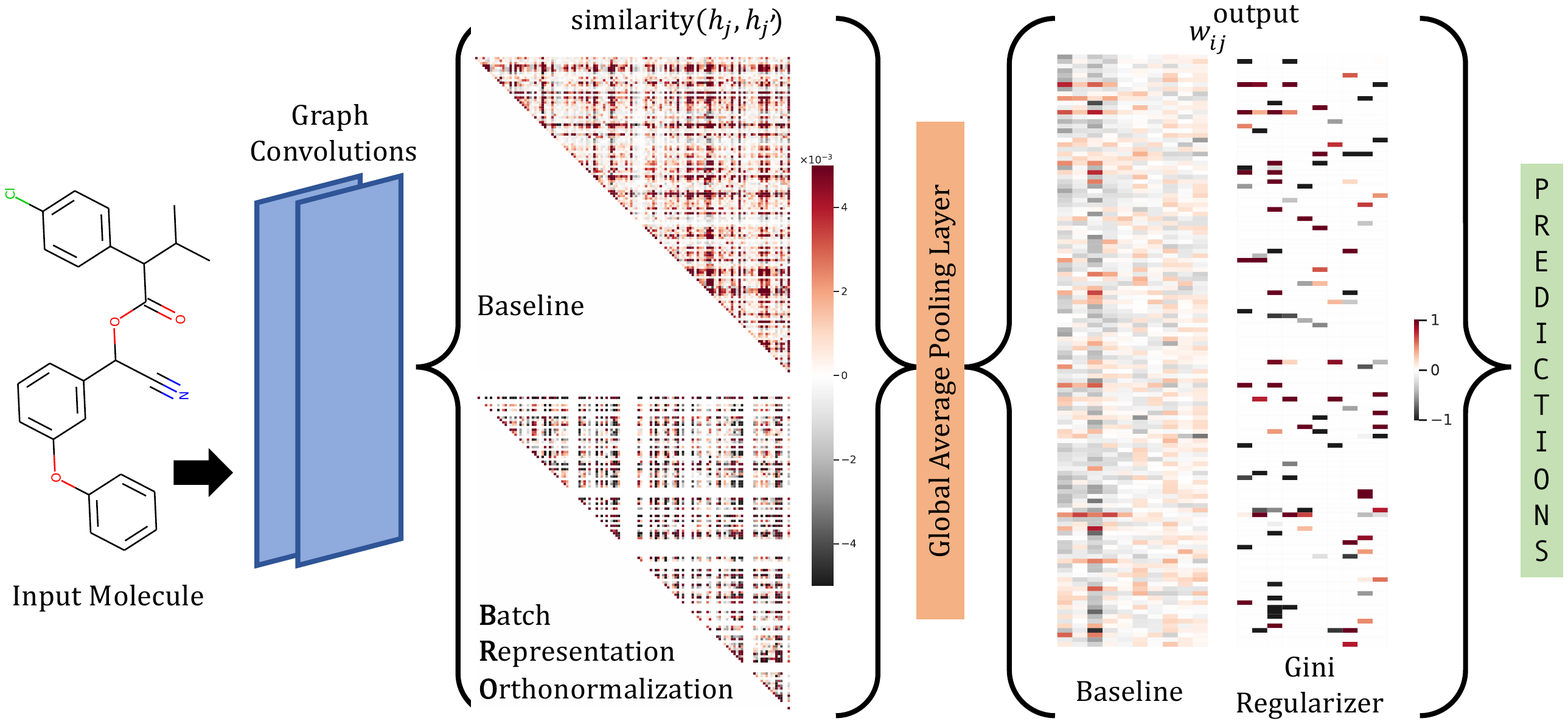}}
        \caption{
        \textbf{Batch Representation Orthnormalization and Gini Regularization}
        From left to right: the input molecule passes through a number of graph convolutional operators, generating new graph embeddings.
        $n$ embeddings of arbitrary dimension (in this figure: 128) are generated, where $n$ is the number of nodes or atoms in the graph.
        During training, a regularization loss will be computed on the final embedding, $\mathbf{H}$, according to the BRO regularizer in Equation~\ref{eq:BRO}.
        The contrast between the cosine similarities among all pairs of node representations $h_j$ for a model trained without (Baseline) and with the BRO regularizer is shown: note that in the BRO case, many more overlaps are pushed to zero.
        The node representations are then aggregated by a global average pooling layer.
        A final linear layer transforms these aggregations into predictions.
        The multi-task model depicted in the figure has 10 outputs.
        The effects of constraining the weights of this layer according to the Gini regularizer (Equation~\ref{eq:gini}) is shown on the right.
        }
        \label{fig:overview}
    \end{center}
    \vskip -0.2in
\end{figure*}

In a GCNN, particular combinations of dimensions of the learned representations can be visually mapped onto the molecule after any convolution.
Existing attribution methods for GCNNs make use of that fact. 
If these dimensions are correlated or if a particular prediction makes use of very many dimensions, the resulting explanations may be confusing and overwhelming for the end user.
To mitigate these problems, we introduce two new methods to improve the interpretability of GCNN predictions: Batch Representation Orthonormalization (BRO) and Gini regularization.
Both are available in the Pytorch Geometric library~\cite{fey_fast_2019}: \url{https://pytorch-geometric.readthedocs.io/en/latest/modules/nn.html#functional}.
An overview of our approach is shown in Figure~\ref{fig:overview}.

\subsection{Batch Representation Orthonormalization (BRO)}

Our work is loosely inspired by linear combination of atomic orbitals (LCAO) theory~\cite{albright_orbital_2013}.
In LCAO theory, a complete set of molecular orbitals $\psi$ is built from the superposition of an orthonormal basis set of atomic electronic wavefunctions: $\psi_j = \sum_i^N c_{ij}\phi_i$ where $\phi_i$ are the electronic atomic basis functions.
The molecular orbitals must also be mutually orthonormal--that is, the coefficients $c_{ij}$ must be normalized so that $\psi_j^* \psi_j = 1$ and $\psi_j^*\psi_{j'} = 0$.

\begin{figure}[t]
    \vskip 0.2in
    \begin{center}
        \centerline{\includegraphics[trim=40 80 320 80, clip, width=\columnwidth]{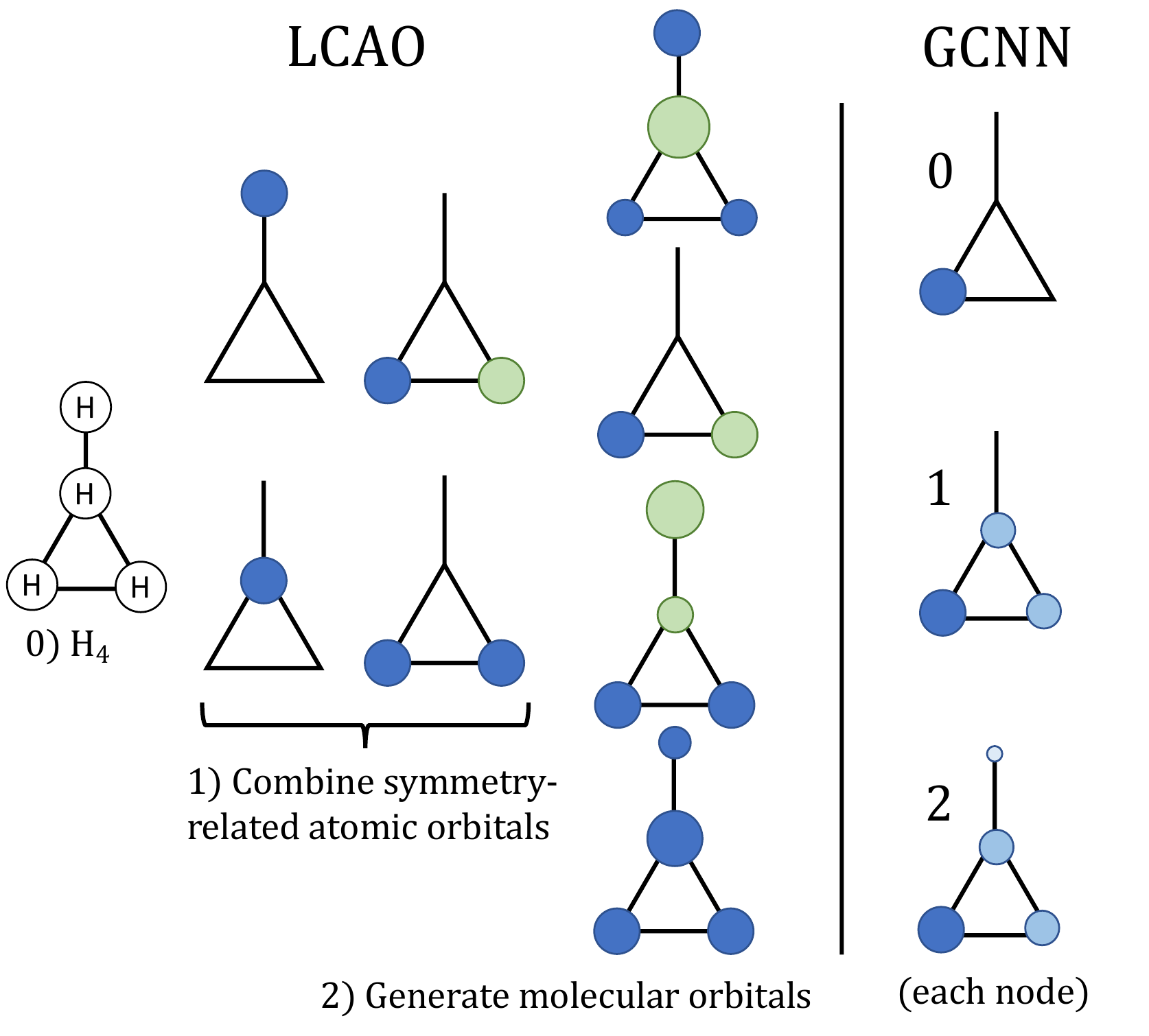}}
        \caption{
        On the left we illustrate the LCAO principle for a hypothetical four-hydrogen molecule ($\sigma$-bonding only).
        The colors represent the sign of the atomic orbital: neighboring atomic sites with different signs are said to be anti-bonding, with correspondingly higher energy.
        Step 0 shows the topology of the molecule.
        Step 1 shows combining atomic orbitals which are symmetry-related in bonding and anti-bonding combinations.
        Step 2 shows a combination of the intermediate orbitals from Step 1 into the final molecular orbitals.
        The atomic contributions in the final molecular orbitals are sized according to the eigenvectors of the adjacency matrix, and they are ranked according to eigenvalue: most bonding (bottom) to most anti-bonding (top).
        The right side of the image shows an analogous process for GCNNs, receiving information from increasingly distant (represented by a lighter shade) neighbors for each convolutional step.
        This happens simultaneously for each node.
        }
        \label{fig:LCAOvsGNN}
    \end{center}
    \vskip -0.2in
\end{figure}

Pictorially, one can think of LCAO as building up the molecular orbitals by sequentially making bonding and anti-bonding combinations of symmetry-related atoms until the full molecular orbital set is constructed.
GCNNs can also be conceptualized as building up a full molecular picture from atomic sites.
Instead of combining symmetry-related sites, GCNNs combine the information from increasingly distant atomic neighbors on the molecule.
This correspondence is not only superficial: the final molecular orbitals obtained by this symmetry mixing procedure can be described by the eigenvectors of the adjacency matrix.
The contrast of these two principles is depicted in Figure~\ref{fig:LCAOvsGNN}.

In the LCAO picture, orthonormality is guaranteed by construction.
The representations $\mathbf{H}$ created by a graph convolution ($h_{ij}$ for the $i$th atom and $j$th molecular embedding; $h_j$ is a vector that can be mapped directly onto the molecule) have no such guarantee.
In this work, we introduce an orthonormality constraint in the training process to partially recover this property.
For a molecule with representation $\mathbf{H}$, the molecular regularization loss is defined by:

\begin{equation}
    \label{eq:BRO}
    \mathcal{L}_{\textrm{BRO}}^\mathrm{mol} = \frac{\lambda}{2
    } \ || \mathbf{HH}^T - \mathbf{I}||_2
\end{equation}

Where $||\cdot||_2$ indicates the vector 2-norm and $\mathbf{I}$ is the identity matrix.
$\lambda$ is a hyperparameter.
Since the forward pass of a GCNN will typically use mini-batching, the regularization loss must be aggregated over all graphs in the batch.
Computing so many normalizations is computationally taxing, so in this work we limit BRO application to the outputs of the final graph convolution.
We observe that training typically takes four times as long.
The effects of this constraint on the node embeddings for a single molecule are shown in the center of Figure~\ref{fig:overview}.

\subsection{Gini Regularization}

Multitask neural networks (MTNNs) and multitask graph convolutional neural networks (MT-GCNNs) are of special interest for modeling molecular properties.
Often, a user is interested in multiple properties of a molecule.
Since many physicochemical properties are related, it makes intuitive sense that internal representations could be reused for different tasks.

A typical MT-GCNN architecture will follow the graph convolutional layers with a global graph aggregation layer, which aggregates the node-level outputs into graph-level outputs for graph property prediction.
A sequential, fully-connected neural network then makes graph level predictions based on the graph-level outputs from the global aggregation layer.
The simplest example of this architecture, which we will use in this paper, is a global average pooling (GAP) layer followed by a single fully-connected layer.

We wish to constrain the weights of the final fully-connected layer to be sparse.
We reason that this should have the two-fold effect of reducing the number of node representations $h_j$ that are relevant to a specific prediction and revealing which representations are shared among tasks.
The usual $\ell_1$ or $\ell_2$ regularization is not appropriate here, as penalizing the magnitude of the weights directly damages the performance of regression metrics.

Instead, we use a regularizer inspired by the Gini coefficient~\cite{gini_variabilita_1912}.\footnote{We first documented this approach in our workshop paper ~\citealt{henderson_gini_2020}.}
The Gini coefficient was invented in the context of economics as a straightforward way to compare income or wealth inequality across different countries~\cite{dorfman_formula_1979}.
A high Gini coefficient implies a inequal wealth distribution, while a low one indicates a more equitable distribution.

For our model, we want the representations to have high inequality: each prediction should be dominated by as few representations as possible.
We select each row $w_i$ of the weights of the final fully-connected layer which is responsible for predicting task $i$ from the $n$ outputs of the GAP layer and compute:

\begin{equation}
    \mathcal{L}_\textrm{Gini}^i = \sum_j^n \sum_{j'}^n \frac{
    | w_{ij} - w_{ij'} |
    } {
    2 (n^2 - n) \bar{w_i}
    }
    \label{eq:gini}
\end{equation}

where $j$ ranges over all weights in the row.
$\bar{w_i}$ is the mean weight value for the row.
The Gini coefficient $\mathcal{L}^i_\textrm{Gini}$ ranges from zero to one: zero if all $w$ are equal and one if one $w_i$ is non-zero and the rest zero.
Since weights of a linear transform are not necessarily restricted to be non-negative, we will always use $|w|$ rather than $w$ for computing the Gini regularization during training.

We take the mean $g = \frac{1}{n}\sum_i^n \mathcal{L}_\textrm{Gini}^i$ over all tasks.
The training loss becomes $L / g^m$ where $L$ is the multi-task regression loss, and $m$ is a hyperparameter to tune the effect of the Gini regularization.
The effects of the Gini regularization on a trained model are shown on the right side of Figure~\ref{fig:overview}.

\section{Related Work}\label{sec:related-work}

Our work is inspired by convolutional neural network (CNN) approaches.
In particular, the work of~\citealt{zhang_interpretable_2018} seeks to align individual convolutional filters to particular concepts, constraining categorical and spatial entropy.
Like our approach, this method requires no annotations.
Forcing orthonormality in the learned representations is conceptually similar to disentanglement or finding a more interpretable ``basis set'': this approach is explored for CNNs in~\citealt{zhou_interpreting_2019}.

Orthogonality or orthormality is often used as a constraint in training neural networks, however this is usually applied to the weights of learnable parameters~\cite{huang_orthogonal_2018}.
Recurrent neural networks in particular stand to benefit from orthonormalization as a means to address exploding or vanishing gradients~\cite{arjovsky_unitary_2016, jing_tunable_2017, vorontsov_orthogonality_2017}.
The exact form of the BRO layer was derived from ~\citealt{xie_all_2017}, although here again applied to learnable parameters rather than learned representations.

In the context of GCNNs, methods such as Graph Information Bottleneck~\cite{wu_graph_2020}, Deep Graph Infomax~\cite{velickovic_deep_2018}, and Infograph~\cite{sun_infograph_2019} seek to constrain representations according to information-theoretic principles.

Many attribution methods for neural networks have been proposed~\cite{zhou_learning_2016, selvaraju_grad-cam_2017, shrikumar_learning_2017, smilkov_smoothgrad_2017, sundararajan_axiomatic_2017} and later adapted to GCNNs ~\cite{xie_interpreting_2019, ying_gnnexplainer_2019, pope_explainability_2019, baldassarre_explainability_2019}.
In this work we focus on Class Activation Maps (CAM)~\cite{zhou_learning_2016}.
CAM is a simple yet powerful method connecting network outputs and the learned node representations.
In GCNNs where the node feature aggregation is done by a GAP layer, CAM corresponds to the scalar product of individual node features by the weights of the output layer:

\begin{equation}
    \textrm{Attribution}(\textrm{atom}_i) = w_j^\mathsf{T}  \cdot h_{i}
    \label{eq:cam}
\end{equation}

where $h_i$ corresponds to the learned feature vector for atom $i$ and $w_j$ corresponds to the row of the output weight matrix of the network corresponding to the $j$th task.
If in the single-task case, $w_j = \mathbf{W}$.
We expect our contributions to specifically help with CAM-like attribution methods: the Gini constraint reduces the number of non-zero elements in $\mathbf{W}$ and the BRO orthonormalization ensures that the different dimensions of the learned atom representations $\mathbf{H}$ are independent from each other.

\section{Experiments and Results}\label{sec:experiments-and-results}

We wish to show the effect of BRO and Gini regularization on model prediction explainability.
To illustrate this, we show the effect of our regularizers on attribution maps on benchmark datasets for which an exact attribution score can be calculated.

Next, we train new models on proprietary assay data to predict physicochemical endpoints.
We generate attribution maps from models trained with and without our constraints and survey experts on which attribution they prefer.

In all experiments, no hyperparmeter tuning was performed: the BRO and Gini regularizers are added to the baseline models with the their respective hyperparameters fixed ($\lambda=0.001$ for BRO regularization and $m=5.0$ for Gini regularization).
We chose values for parameters $\lambda$ and $m$ qualitatively: we sought the largest values for each that did not degrade evaluation metrics too much (see Section~\ref{subsec:model-performance}).

\subsection{Attribution Benchmarks}\label{subsec:attribution-benchmarks}

Recently,~\citealt{sanchez-lengeling_evaluating_2020} proposed three artificial tasks to evaluate attribution methods on molecular GCNNs.
These simple tasks have an associated ground truth explanation that can be used to score any attribution method's output.
For instance, if we trained a model to predict whether a molecule contains a benzene ring or not, a perfect attribution would highlight only the atoms in the benzene ring and no others.
We followed their protocol and metrics to benchmark the effect of our proposed regularizers on the attribution performance.

We focus on CAM and CAM-derived attribution methods, including GradCAM~\cite{selvaraju_grad-cam_2017} either on all convolutional layers or only the last one.
This is because the authors found CAM to systematically outperform other attribution methods.
Since our regularization method has not been implemented yet on edge features, we focus on the benchmark graph architectures that use node features exclusively (graph convolutional network (GCN) and graph attention network (GAT) in~\citealt{sanchez-lengeling_evaluating_2020}).
We also introduce a variation on CAM that we call TopRep.
TopRep applies the CAM Equation~\ref{eq:cam} using only the weight of the output layer that has the highest magnitude: that is, $w_j$ of Equation~\ref{eq:cam} retains the entry with the largest absolute value, and all others are set to zero.

\subsubsection{Dataset and Network Architecture}

\citealt{sanchez-lengeling_evaluating_2020} present three attribution tasks: ``Benzene'', ``Amine-Ether-Benzene'' and ``CrippenLogP.''
The Benzene task is as described above.
The Amine-Ether-Benzene task tests if each of the Amine, Ether, and Benzene fragments are present in the molecule (logical AND).
Both of these attribution tasks are scored using AUROC~\cite{bradley_use_1997}.
The underlying classification task is also scored with AUROC.
The CrippenLogP task is a regression task, with both attribution and regression scored with Pearson correlation.
We discuss this task in detail at the end of this section.

For the GCN and GAT models, we use the hyperparemeters given in Sanchez-Lengeling and modify their code only insofar as necessary to add the BRO and Gini regularizers.
Eighty trials of each combination are run.
Our modifications to the benchmarking code are available at \url{https://github.com/bayer-science-for-a-better-life/graph-attribution}.

\subsubsection{Results}

\begin{figure}
    \centering
    \begin{subfigure}{\linewidth}
        \centering
        \includegraphics[trim=40 85 340 83, clip, width=1.0\linewidth]{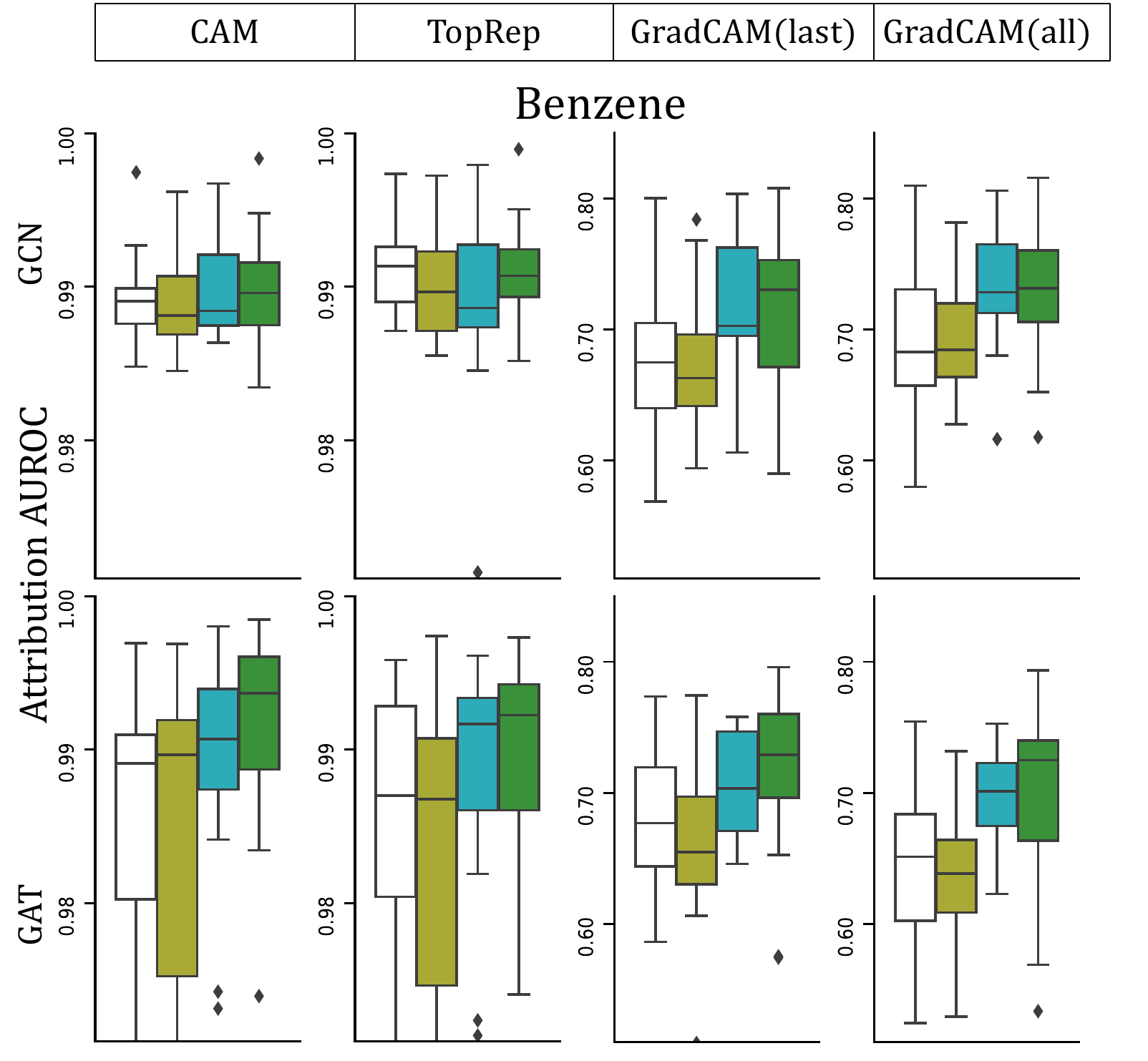}
    \end{subfigure}
    \begin{subfigure}{\linewidth}
        \centering
        \includegraphics[trim=40 85 340 113, clip, width=1.0\linewidth]{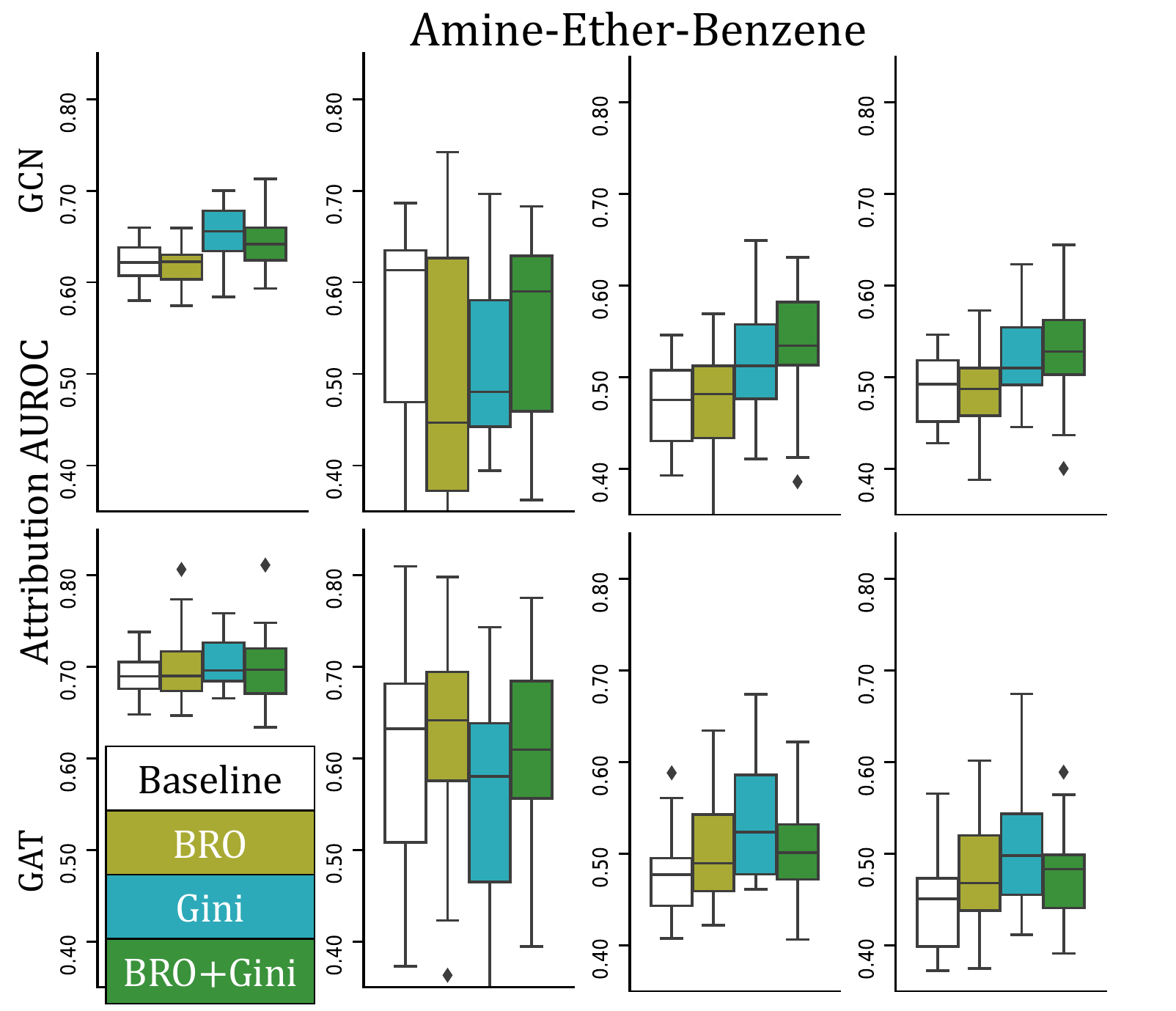}
    \end{subfigure}
    \begin{subfigure}{\linewidth}
        \centering
        \includegraphics[trim=40 90 340 113, clip, width=1.0\linewidth]{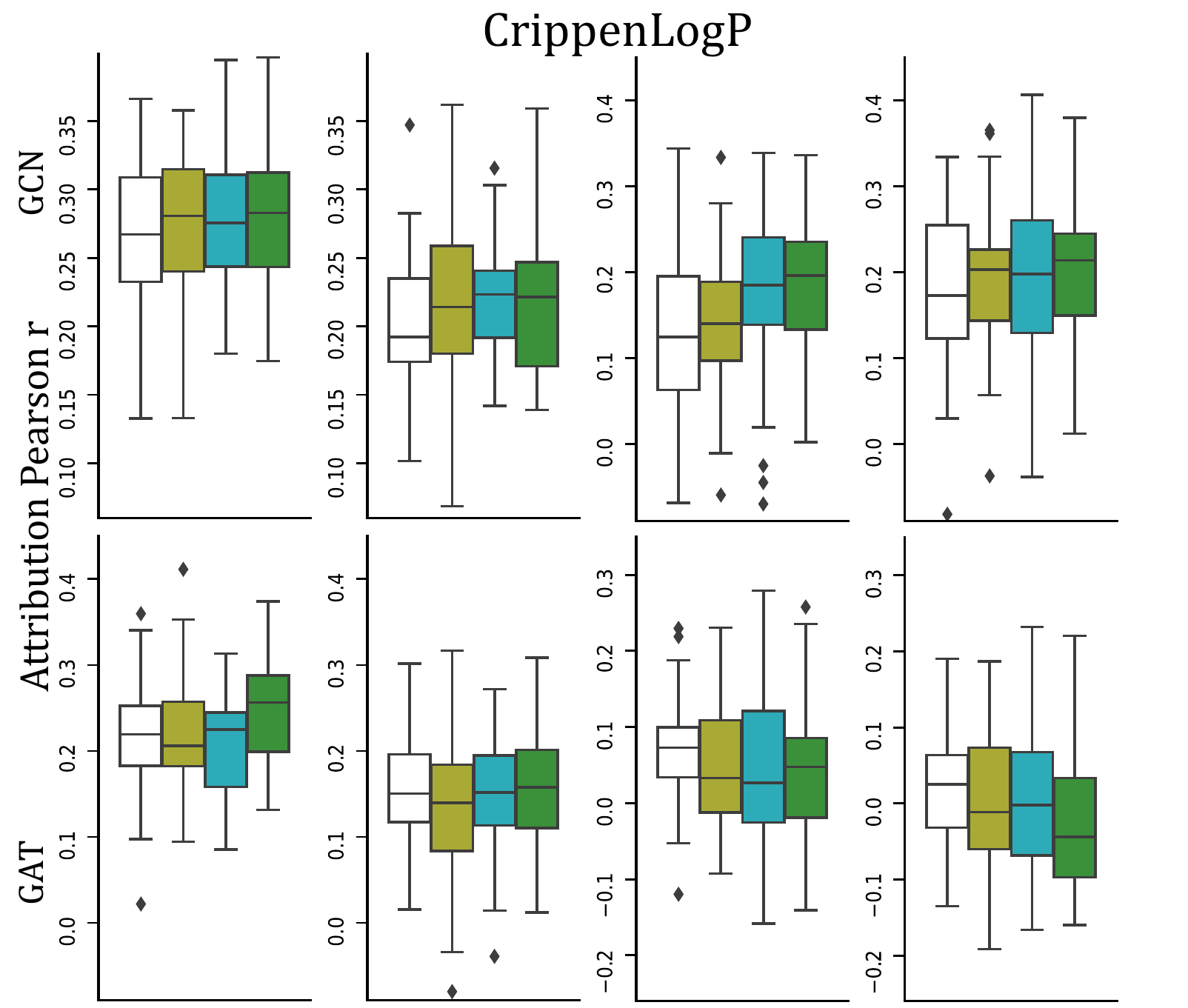}
    \end{subfigure}
    \caption{
    Boxplots of attribution AUROC scores for all tasks.
    Both GCN and GAT models were considered.
    }
    \label{fig:attribution_boxplots}
\end{figure}

Figure~\ref{fig:attribution_boxplots} show the detailed distributions of attribution scores for the Benzene, Amine-Ether-Benzene, and CrippenLogP tasks ($n=80$ for every model/task combination).
For many combinations, BRO or Gini regularization markedly improve the attribution AUROC score.
Often, the combination of BRO and Gini yields better results than either regularizer alone.
This effect can be most strongly seen in the Benzene GAT combination (bottom row).
This bolsters our reasoning that it takes both sparsity \textit{and} orthogonal respresentations to generate a good attribution.
While there are some task/model combinations for which none of our constraints improve attribution, no combination's attribution AUROC score suffers significantly from the Gini+BRO regularizers.

TopRep also gives competitive results, even for the baseline configuration.
This is surprising in the baseline case, because TopRep is implicitly throwing out a lot of information if the output weight matrix has not been made sparse through, for example, Gini regularization.

Table~\ref{tab:CAM_benchmark} summarizes our results on all three benchmark tasks, including CrippenLogP.

\begin{table}[h]
    \scriptsize
    \centering
    \caption{Performance of the attribution methods on the three benchmark tasks (mean for 80 trials per combination).
    For classification tasks, we report the attribution AUROC score and for regression the Pearson correlation.
    Best value in each column in bold.
    }
    \label{tab:CAM_benchmark}
    \resizebox{\columnwidth}{!}{
    \begin{tabular}{llrrrrrr}
        \toprule
        &       & \multicolumn{2}{c}{Benzene} & \multicolumn{2}{c}{AmEthBenz} & \multicolumn{2}{c}{CrippenLogP} \\
        \multicolumn{1}{l}{Attr.} & Const. & GCN   & GAT   & GCN   & GAT   & GCN   & GAT \\
        \midrule
        Random & None  & 0.614  & 0.613  & \cellcolor[rgb]{ .914,  .949,  .886} 0.500 & {\cellcolor[rgb]{ .898,  .941,  .867}} 0.495 & {\cellcolor[rgb]{ 1,  .996,  .992}} -0.086 & -0.093 \\
        & BRO   & 0.613  & 0.613  & \cellcolor[rgb]{ .894,  .937,  .863} 0.507 & \cellcolor[rgb]{ .878,  .933,  .843} 0.503 & -0.092 & -0.093 \\
        & Gini  & \cellcolor[rgb]{ .996,  1,  1} 0.614 & 0.613  & \cellcolor[rgb]{ .922,  .957,  .898} 0.498 & \cellcolor[rgb]{ .886,  .937,  .855} 0.500 & -0.091 & \cellcolor[rgb]{ 1,  .996,  .992} -0.088 \\
        & both  & 0.612  & 0.615  & \cellcolor[rgb]{ .918,  .953,  .89} 0.499 & \cellcolor[rgb]{ .886,  .937,  .855} 0.500 & -0.091 & \cellcolor[rgb]{ 1,  1,  .996} -0.091 \\
        \midrule
        Grad & None  & \cellcolor[rgb]{ .875,  .906,  .961} 0.678 & \cellcolor[rgb]{ .89,  .918,  .969} 0.670 & \cellcolor[rgb]{ .988,  .996,  .984} 0.475 & \cellcolor[rgb]{ .941,  .965,  .922} 0.476 & \cellcolor[rgb]{ .961,  .698,  .518} 0.130 & \cellcolor[rgb]{ .969,  .757,  .612} 0.071 \\
        CAM & BRO   & \cellcolor[rgb]{ .882,  .914,  .965} 0.673 & \cellcolor[rgb]{ .91,  .933,  .973} 0.659 & 0.471  & \cellcolor[rgb]{ .886,  .937,  .855} 0.500 & \cellcolor[rgb]{ .957,  .682,  .494} 0.141 & \cellcolor[rgb]{ .973,  .792,  .667} 0.048 \\
        (last) & Gini  & \cellcolor[rgb]{ .796,  .847,  .937} 0.718 & \cellcolor[rgb]{ .835,  .878,  .949} 0.697 & \cellcolor[rgb]{ .851,  .918,  .812} 0.520 & \cellcolor[rgb]{ .808,  .89,  .753} 0.536 & \cellcolor[rgb]{ .949,  .635,  .42} 0.176 & \cellcolor[rgb]{ .973,  .796,  .675} 0.045 \\
        & both  & \cellcolor[rgb]{ .8,  .851,  .937} 0.715 & \cellcolor[rgb]{ .8,  .851,  .937} 0.716 & \cellcolor[rgb]{ .8,  .886,  .741} 0.537 & \cellcolor[rgb]{ .882,  .933,  .847} 0.503 & \cellcolor[rgb]{ .949,  .627,  .404} 0.182 & \cellcolor[rgb]{ .973,  .8,  .682} 0.041 \\
        \midrule
        Grad & None  & \cellcolor[rgb]{ .847,  .886,  .953} 0.692 & \cellcolor[rgb]{ .973,  .98,  .992} 0.628 & \cellcolor[rgb]{ .949,  .973,  .933} 0.488 & 0.448  & \cellcolor[rgb]{ .949,  .627,  .408} 0.181 & \cellcolor[rgb]{ .976,  .827,  .729} 0.023 \\
        CAM & BRO   & \cellcolor[rgb]{ .851,  .886,  .953} 0.691 & \cellcolor[rgb]{ .965,  .973,  .988} 0.632 & \cellcolor[rgb]{ .969,  .984,  .961} 0.482 & \cellcolor[rgb]{ .937,  .965,  .918} 0.477 & \cellcolor[rgb]{ .949,  .608,  .376} 0.194 & \cellcolor[rgb]{ .98,  .859,  .773} 0.003 \\
        (all) & Gini  & \cellcolor[rgb]{ .769,  .824,  .929} 0.733 & \cellcolor[rgb]{ .855,  .894,  .957} 0.687 & \cellcolor[rgb]{ .843,  .91,  .796} 0.523 & \cellcolor[rgb]{ .878,  .929,  .843} 0.505 & \cellcolor[rgb]{ .949,  .612,  .384} 0.192 & \cellcolor[rgb]{ .984,  .867,  .784} -0.002 \\
        & both  & \cellcolor[rgb]{ .769,  .827,  .929} 0.732 & \cellcolor[rgb]{ .835,  .875,  .949} 0.698 & \cellcolor[rgb]{ .824,  .898,  .769} 0.530 & \cellcolor[rgb]{ .945,  .969,  .925} 0.474 & \cellcolor[rgb]{ .945,  .6,  .365} 0.201 & \cellcolor[rgb]{ .988,  .894,  .835} -0.022 \\
        \midrule
        CAM   & None  & \cellcolor[rgb]{ .275,  .451,  .773} \textcolor[rgb]{ 1,  1,  1}{0.989} & \cellcolor[rgb]{ .278,  .459,  .773} \textcolor[rgb]{ 1,  1,  1}{0.984} & \cellcolor[rgb]{ .537,  .737,  .404} \textcolor[rgb]{ 1,  1,  1}{0.621} & \cellcolor[rgb]{ .467,  .694,  .314} \textcolor[rgb]{ 1,  1,  1}{0.691} & \cellcolor[rgb]{ .933,  .506,  .216} \textcolor[rgb]{ 1,  1,  1}{0.268} & \cellcolor[rgb]{ .937,  .533,  .259} \textcolor[rgb]{ 1,  1,  1}{0.221} \\
        & BRO   & \cellcolor[rgb]{ .275,  .451,  .773} \textcolor[rgb]{ 1,  1,  1}{0.989} & \cellcolor[rgb]{ .29,  .467,  .776} \textcolor[rgb]{ 1,  1,  1}{0.977} & \cellcolor[rgb]{ .545,  .741,  .412} \textcolor[rgb]{ 1,  1,  1}{0.619} & \cellcolor[rgb]{ .447,  .682,  .29} \textcolor[rgb]{ 1,  1,  1}{0.701} & \cellcolor[rgb]{ .933,  .498,  .204} \textcolor[rgb]{ 1,  1,  1}{0.274} & \cellcolor[rgb]{ .937,  .537,  .263} \textcolor[rgb]{ 1,  1,  1}{0.219} \\
        & Gini  & \cellcolor[rgb]{ .271,  .451,  .773} \textcolor[rgb]{ 1,  1,  1}{0.990} & \cellcolor[rgb]{ .267,  .447,  .769} \textcolor[rgb]{ 1,  1,  1}{\textbf{0.989}} & \cellcolor[rgb]{ .439,  .678,  .278} \textcolor[rgb]{ 1,  1,  1}{\textbf{0.653}} & \cellcolor[rgb]{ .439,  .678,  .278} \textcolor[rgb]{ 1,  1,  1}{\textbf{0.703}} & \cellcolor[rgb]{ .933,  .494,  .196} \textcolor[rgb]{ 1,  1,  1}{0.278} & \cellcolor[rgb]{ .941,  .553,  .29} \textcolor[rgb]{ 1,  1,  1}{0.208} \\
        & both  & \cellcolor[rgb]{ .271,  .451,  .773} \textcolor[rgb]{ 1,  1,  1}{0.990} & \cellcolor[rgb]{ .275,  .451,  .773} \textcolor[rgb]{ 1,  1,  1}{\textbf{0.989}} & \cellcolor[rgb]{ .463,  .69,  .306} \textcolor[rgb]{ 1,  1,  1}{0.646} & \cellcolor[rgb]{ .451,  .686,  .294} \textcolor[rgb]{ 1,  1,  1}{0.699} & \cellcolor[rgb]{ .929,  .49,  .192} \textcolor[rgb]{ 1,  1,  1}{\textbf{0.279}} & \cellcolor[rgb]{ .929,  .49,  .192} \textcolor[rgb]{ 1,  1,  1}{\textbf{0.248}} \\
        \midrule
        TopRep & None  & \cellcolor[rgb]{ .267,  .447,  .769} \textcolor[rgb]{ 1,  1,  1}{\textbf{0.991}} & \cellcolor[rgb]{ .286,  .463,  .776} \textcolor[rgb]{ 1,  1,  1}{0.979} & \cellcolor[rgb]{ .737,  .851,  .663} 0.556 & \cellcolor[rgb]{ .678,  .816,  .584} 0.595 & \cellcolor[rgb]{ .945,  .592,  .353} 0.205 & \cellcolor[rgb]{ .953,  .635,  .42} 0.153 \\
        & BRO   & \cellcolor[rgb]{ .286,  .463,  .776} \textcolor[rgb]{ 1,  1,  1}{0.982} & \cellcolor[rgb]{ .314,  .482,  .784} \textcolor[rgb]{ 1,  1,  1}{0.967} & \cellcolor[rgb]{ .957,  .976,  .945} 0.486 & \cellcolor[rgb]{ .627,  .788,  .518} 0.619 & \cellcolor[rgb]{ .945,  .58,  .337} 0.214 & \cellcolor[rgb]{ .957,  .667,  .471} 0.131 \\
        & Gini  & \cellcolor[rgb]{ .282,  .459,  .776} \textcolor[rgb]{ 1,  1,  1}{0.984} & \cellcolor[rgb]{ .271,  .451,  .773} \textcolor[rgb]{ 1,  1,  1}{0.988} & \cellcolor[rgb]{ .882,  .933,  .847} 0.510 & \cellcolor[rgb]{ .761,  .863,  .69} 0.558 & \cellcolor[rgb]{ .945,  .588,  .349} 0.208 & \cellcolor[rgb]{ .953,  .643,  .431} 0.148 \\
        & both  & \cellcolor[rgb]{ .271,  .451,  .773} \textcolor[rgb]{ 1,  1,  1}{\textbf{0.991}} & \cellcolor[rgb]{ .271,  .451,  .773} \textcolor[rgb]{ 1,  1,  1}{0.987} & \cellcolor[rgb]{ .741,  .855,  .667} 0.555 & \cellcolor[rgb]{ .682,  .82,  .592} 0.593 & \cellcolor[rgb]{ .945,  .596,  .357} 0.204 & \cellcolor[rgb]{ .949,  .631,  .416} 0.155 \\
        \bottomrule
    \end{tabular}%
    }
    \label{tab:addlabel}%
\end{table}%

We notice that the CAM-related attribution methods work poorly for the regression task CrippenLogP.
This was already observed in~\citealt{sanchez-lengeling_evaluating_2020}, where only the integrated gradients approach~\cite{sundararajan_axiomatic_2017} seemed to perform slightly better than other attribution methods.

We interpret this finding in the construction of the benchmark task itself.
The authors used a dataset with experimentally determined solubility~\cite{delaney_esol_2004} and consider as ground truth explanation the atom contribution for the water-octanol partition coefficient as calculated by the Crippen LogP empirical model~\cite{wildman_prediction_1999}.
Although logP and solubility are correlated, we assume that logP is not always a good explanation for a more complex endpoint like solubility, and therefore it should not be surprising if the built model's explanations do not align very well with the logP atom contributions.
A better task might be to build a model for logP and use the Crippen LogP attributions as ground truth explanations.
This experiment was not performed here to stay in the framework of the benchmark.

\subsection{Physicochemical endpoints}\label{subsec:physicochemical-tasks}

The results on the constructed benchmark dataset in the previous section encouraged us to evaluate our methods against expert opinion: a much costlier undertaking.

It is impossible to reliably and quickly calculate properties like solubility or binding affinity $ab\ initio$ from the chemical structure of a compound, let alone attribute the prediction to specific sites or functional groups.
When screening drug candidates, a medicinal chemist may synthesize many similar molecules with the desired activity while simultaneously trying to minimize or maximize some other property.

Thus, being able to guess which modifications may alter these properties is extremely helpful.
Here, we seek to mimic the chemists' intuition, and see if the BRO and Gini constraints may produce models that generate attributions that more closely match their instincts.

\subsubsection{Dataset}

The data used for building the multitask GCNN model is a dataset of measured physicochemical properties for small molecules from 10 different assays.
Extensive discussion of the preparation and characteristics of the dataset can be found in our previous work~\citealt{montanari_modeling_2019}.
There are a total of 537,443 compounds with assays covering properties such as solubility, lipophilicity (logD at two different pHs), melting point, human serum albumin binding and membrane affinity; 79\% of the compounds are measured for only one endpoint, 11\% for two, 9\% for 3, and 1\% four or more.

\begin{figure}
    \vskip 0.2in
    \begin{center}
        \centerline{\includegraphics[trim=40 80 250 80, clip, width=\columnwidth]{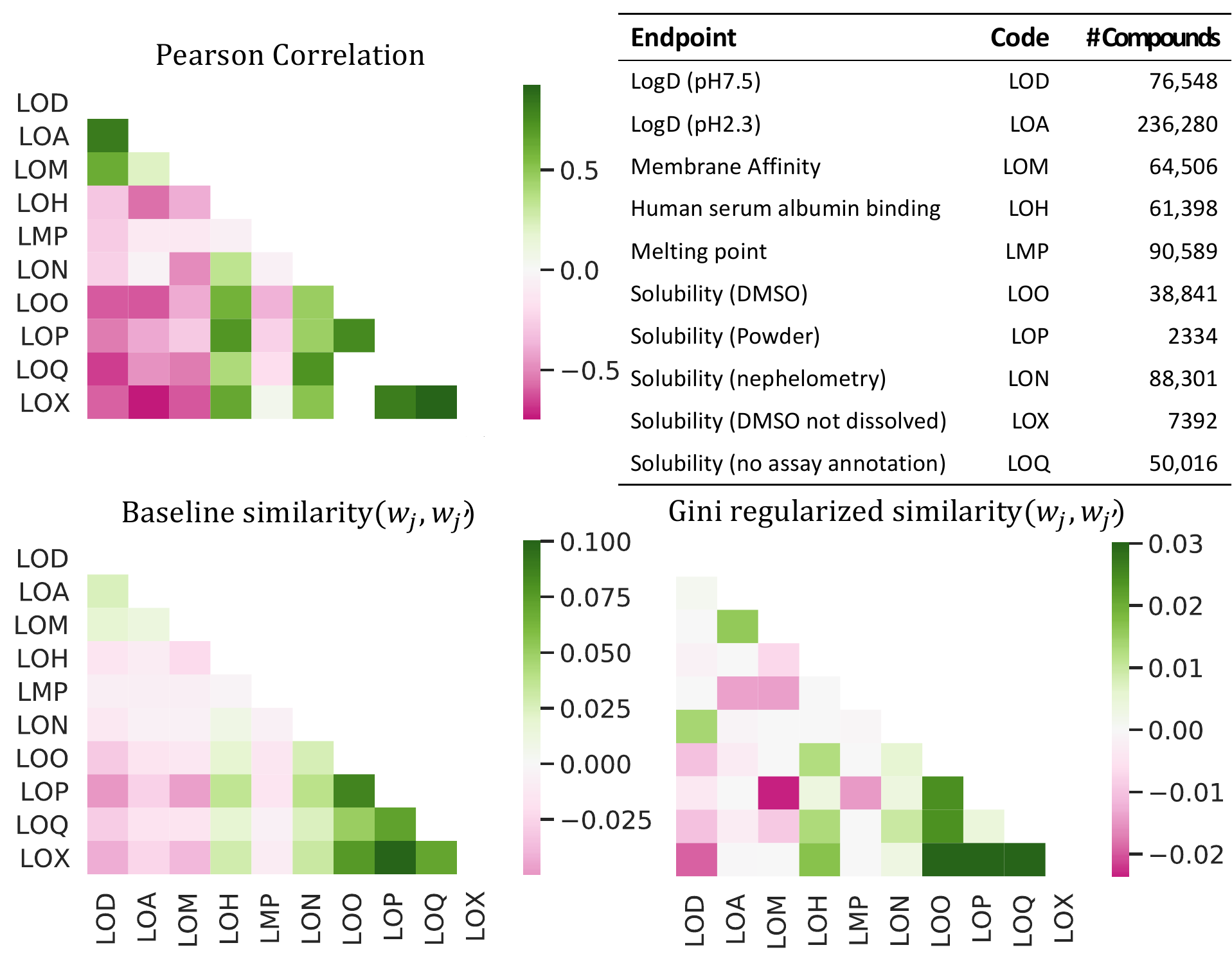}}
        \caption{
        The first row shows the Pearson correlation between each pair of endpoints, along with a table giving a description and count of each.
        The second row shows the cosine similarity of the rows of an output weight matrix for an unconstrained and Gini-constrained model.
        }
        \label{fig:gini_correlations}
    \end{center}
    \vskip -0.2in
\end{figure}

All ten endpoints along with their frequencies and the correlation between pairs is shown in Figure~\ref{fig:gini_correlations}.
We also show the cosine similarities of the rows of the trained output weights corresponding to each endpoint.
The similarities of both the baseline and Gini-constrained model mimic the underlying correlation between the measurements, but the Gini-constrained version is sparser.
This may imply it has learned a more specific relationship between the endpoints than merely the data distribution.

Seven cross-validation folds were generated using $k$-means clustering on the ECFC6 fingerprints~\cite{rogers_extended-connectivity_2010}.

\subsubsection{Network Architecture}

The model, initially built in Tensorflow with the DeepChem library ~\cite{ramsundar_deep_2019} was converted to Pytorch Geometric for easier access and customization~\cite{fey_fast_2019}.
We tried as much as possible to follow the DeepChem graph convolution mechanism which is an implementation of the~\citealt{duvenaud_convolutional_2015} algorithm.

Input node features are computed with DeepChem and consist of 75 atomic properties: number of radical electrons, whether the atom is aromatic, and one-hot encoding on each of the atom type, degree, formal charge, and hybridization.
Two convolutional layers are then applied, then a GAP layer going from individual node features to a global graph encoding.
A hyperbolic tangent function is applied to the graph encoding before the final linear layer.

Our models were trained using the ADAM optimizer~\cite{kingma_adam_2017} for forty epochs with exponential learning rate decay with a base of 0.97 decaying every 1000 steps.
No hyperparemeter search is performed: models with either or both of BRO and Gini regularization use the same hyperparameters as the baseline models.
Model code is available in the Supplementary Material.

\subsubsection{Survey of Medicinal Chemists}

To test our methods in a realistic setting, we prepared a survey to gather opinions from human experts.
We collected public compounds with reported measured solubility~\cite{sorkun_aqsoldb_2019}, logD~\cite{alelyunas_high_2010, low_optimised_2016} or melting point~\cite{williams_melting_2015}.
These compounds were chosen for their problematic properties: they were either very insoluble, having a high logD, or a low melting point.

We then used the multitask GCN built on physicochemical data (in both the constrained and non constrained version) to predict the solubility, logD or melting point of those molecules.
Because the data is collected from public sources, potentially many different assays were used to measure the different endpoints of interest. 
Therefore, an exact match of the predictive model (built on proprietary established assay data) with the publicly reported values is not expected. 
Since we focused on problematic molecules, we compared predictions and reported experimental values qualitatively only. 
For example, insoluble molecules with a predicted solubility over 30 mg/L were discarded.
In general, molecules with grossly inaccurate predictions by the models were excluded from the survey.
In total, 10 molecules were kept for solubility, 8 for logD and 6 for melting point.
Attribution methods were then applied on each predicted molecule: either CAM (our baseline, using the unconstrained model), CAM$_{\mathrm{BRO+Gini}}$ (CAM applied to the Gini-sparsified and BRO-disentangled model), TopRep$_{\mathrm{BRO+Gini}}$ (CAM using only the top weight from the constrained model output), or a random attribution map (any node representation generated by the constrained model, selected randomly).

\begin{figure}[ht]
    \vskip 0.2in
    \begin{center}
        \centerline{\includegraphics[trim=340 170 340 200, clip, width=\columnwidth]{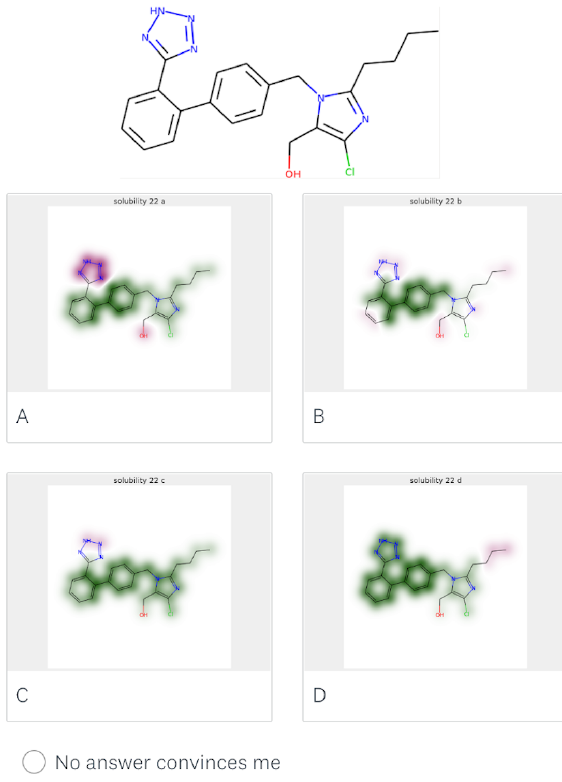}}
        \caption{
        An example question from the survey of medicinal chemists: ``Which atoms drive the low solubility of the molecule? (Green: lower solubility / Magenta: Higher solubility.''
        This shows one of the query molecules for solubility taken from~\cite{sorkun_aqsoldb_2019}  with various site attributions for solubility (LOO task).
        Answers were randomly ordered for each question, blind to both us and the medicinal chemists.
        In this case we have a) CAM$_{\mathrm{BRO+Gini}}$ b) random c) CAM (baseline) d) TopRep$_{\mathrm{BRO+Gini}}$.
        }
        \label{fig:medchemq}
    \end{center}
    \vskip -0.2in
\end{figure}

\begin{figure}
    \vskip 0.2in
    \begin{center}
        \centerline{\includegraphics[trim=40 80 320 80, clip, width=\columnwidth]{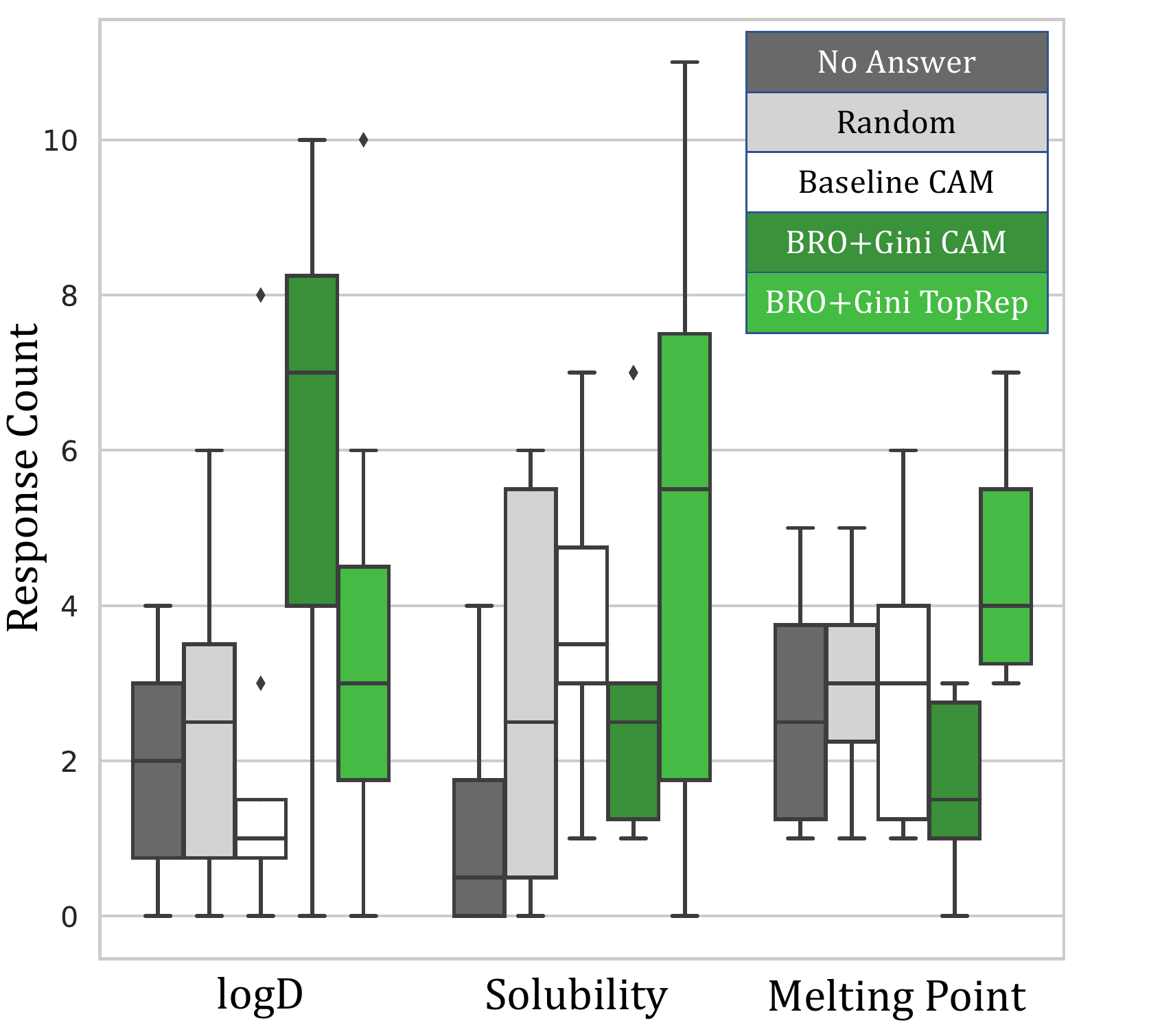}}
        \caption{
        Results of survey of medicinal chemists.
        Not all participants answered all questions.
        The plot shows the distribution of responses for each option for each endpoint.
        For example: for logD questions, a median of 7 respondents (ranging from 0 to 10) picked CAM$_{\mathrm{BRO+Gini}}$, with 50\% of the logD questions receiving between 4 to 8 votes for CAM$_{\mathrm{BRO+Gini}}$.
        }
        \label{fig:survey}
    \end{center}
    \vskip -0.2in
\end{figure}

The users always had the possibility to skip a question or select a negative answer (``No answer convinces me'').
The attribution maps were plotted on the molecule structure using the RDKit's SimilarityMap functionality~\cite{landrum_rdkit_2006}.
An example question as shown to the medicinal chemists is shown in Figure~\ref{fig:medchemq}.
15 medicinal chemists answered the 24 questions and the results are shown in Figure~\ref{fig:survey}. 
The participants were all employed at Bayer on two different sites in Germany, and were at different stages of their career.
They had no previous explicit experience with such specific molecules or tasks although the molecules are known drugs and the tasks are tasks they reason about on a daily basis.

CAM$_{\mathrm{BRO+Gini}}$ and TopRep$_{\mathrm{BRO+Gini}}$ are the most voted attribution methods among the 24 questions, with CAM$_{\mathrm{BRO+Gini}}$ being favored for logD and TopRep$_{\mathrm{BRO+Gini}}$ for solubility and melting point.
Across all endpoints, they are both picked more often than would be expected at random (binomial test, p-value=0.01 and p-value $<$ 0.001 respectively).
Surprisingly, Baseline CAM cannot be distinguished from the Random attribution result (binomial test, p-value=0.07).

This real-life experiment shows that, with a predictive model that is in practice useful for medicinal chemists, the regularization constraints (Gini and BRO) lead to significantly preferred attribution maps.
We also note that the logD endpoint is the easiest for chemists to rationalize and led to the most imbalanced votes in favor of CAM$_{\mathrm{BRO+Gini}}$ (coefficient of variation $\sigma/\mu=0.95$).
Solubility and melting point are less straightforward, and, accordingly, chemists tend to disagree more in their voting ($\sigma/\mu=0.86$ and 0.59 respectively).

All survey questions with attribution maps, including aggregated responses and answer key, are available in the Supplementary Material.

\subsection{Model Performance}\label{subsec:model-performance}

\begin{figure}
    \vskip 0.2in
    \begin{center}
        \centerline{\includegraphics[trim=35 225 390 80, clip, width=\columnwidth]{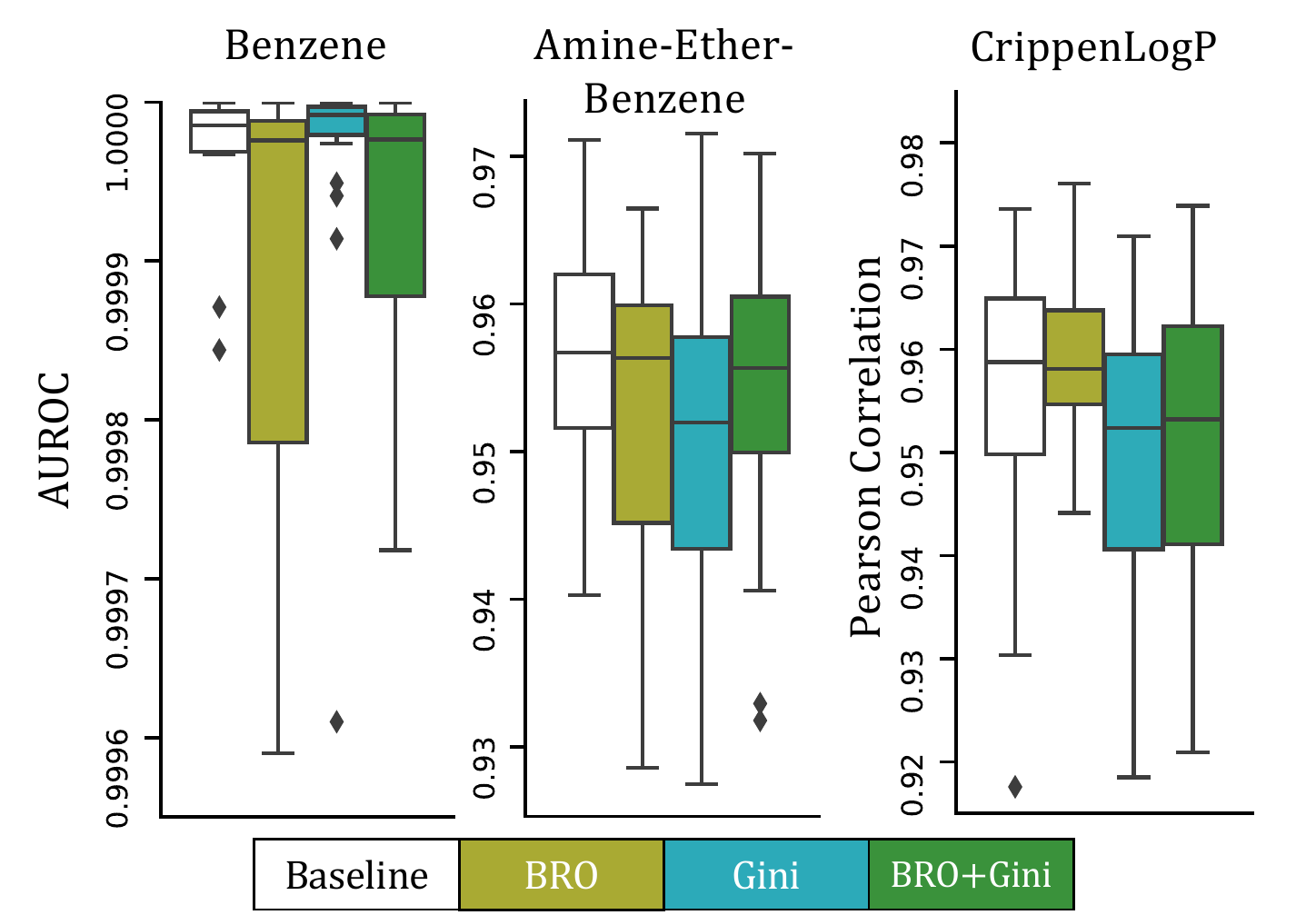}}
        \caption{
        Effect of regularization techniques on the evaluation metric for the Benzene, Amine-Ether-Benzene, and CrippenLogP tasks.
        }
        \label{fig:auroc}
    \end{center}
    \vskip -0.2in
\end{figure}

The Gini and BRO regularizations have a slight negative impact on the classification or regression metrics of the models.
The impact on the AUROC scores for the ``Benzene'' and ``Amine-Ether-Benzene'' classification tasks and the Pearson correlation for the ``CrippenLogP'' regression task described in Section~\ref{subsec:attribution-benchmarks} are illustrated in Figure~\ref{fig:auroc}.

As with the Attribution AUROC, it is nearly impossible to improve the classification AUROC of the Benzene task: all models are extremely close to perfect classification.
Even in this regime, it is clear that the BRO layer negatively effects performance.
In the Amine-Ether-Benzene task, the differences are not significant.
This may be because both BRO and Gini regularization were designed with multitask models in mind, and the Amine-Ether-Benzene is really a multitask problem in disguise.
Finally, adding the BRO and Gini constraints to the CrippenLogP task does not drastically alter model performance, despite doing little to improve interpretability.

The various endpoints of the multitask models trained on physicochemical data described in Section~\ref{subsec:physicochemical-tasks} are affected unevenly.
The cross-validation $R^2$ scores are shown in Table~\ref{tab:admet_r2}.
The solubility endpoints LOP and LOX are strongly affected.
These are the two least represented assays in the dataset, with 2,334 and 7,392 out of 537,443 compounds, respectively.

\begin{table}[htb]
  \centering
  \small
  \caption{Effect of BRO and Gini regularizers on predictive performance for all physicochemical endpoints.
  Mean of scaffold-split cross-validation $R^2$ score and standard deviation in parenthesis.}
  \label{tab:admet_r2}
    \resizebox{\columnwidth}{!}{
    \begin{tabular}{lrrrr}
    \toprule
          & Baseline & BRO   & Gini  & BRO + Gini \\
    \midrule
    LOD   & \cellcolor[rgb]{ .647,  .647,  .647} 0.85 (0.03) & \cellcolor[rgb]{ .682,  .686,  .255} 0.82 (0.03) & \cellcolor[rgb]{ .212,  .686,  .733} \textcolor[rgb]{ 1,  1,  1}{0.83 (0.03)} & \cellcolor[rgb]{ .29,  .6,  .29} \textcolor[rgb]{ 1,  1,  1}{0.81 (0.04)} \\
    LOA   & \cellcolor[rgb]{ .667,  .667,  .667} 0.82 (0.04) & \cellcolor[rgb]{ .694,  .698,  .282} 0.80 (0.04) & \cellcolor[rgb]{ .227,  .69,  .737} \textcolor[rgb]{ 1,  1,  1}{0.82 (0.04)} & \cellcolor[rgb]{ .318,  .616,  .318} \textcolor[rgb]{ 1,  1,  1}{0.79 (0.04)} \\
    LOM   & \cellcolor[rgb]{ .835,  .835,  .835} 0.56 (0.13) & \cellcolor[rgb]{ .863,  .863,  .671} 0.53 (0.13) & \cellcolor[rgb]{ .659,  .863,  .886} 0.53 (0.12) & \cellcolor[rgb]{ .71,  .835,  .71} 0.51 (0.13) \\
    LOH   & \cellcolor[rgb]{ .875,  .875,  .875} 0.50 (0.07) & \cellcolor[rgb]{ .89,  .894,  .741} 0.48 (0.09) & \cellcolor[rgb]{ .749,  .902,  .918} 0.47 (0.09) & \cellcolor[rgb]{ .808,  .89,  .808} 0.44 (0.10) \\
    LMP   & \cellcolor[rgb]{ .914,  .914,  .914} 0.44 (0.08) & \cellcolor[rgb]{ .91,  .91,  .784} 0.45 (0.08) & \cellcolor[rgb]{ .792,  .918,  .929} 0.44 (0.08) & \cellcolor[rgb]{ .82,  .898,  .82} 0.43 (0.09) \\
    LON   & \cellcolor[rgb]{ .867,  .867,  .867} 0.51 (0.09) & \cellcolor[rgb]{ .89,  .894,  .741} 0.48 (0.09) & \cellcolor[rgb]{ .718,  .886,  .906} 0.49 (0.08) & \cellcolor[rgb]{ .792,  .882,  .792} 0.45 (0.10) \\
    LOO   & \cellcolor[rgb]{ .875,  .875,  .875} 0.50 (0.20) & \cellcolor[rgb]{ .902,  .906,  .773} 0.46 (0.24) & \cellcolor[rgb]{ .749,  .902,  .918} 0.47 (0.20) & \cellcolor[rgb]{ .792,  .882,  .792} 0.45 (0.23) \\
    LOP   & \cellcolor[rgb]{ .894,  .894,  .894} 0.47 (0.15) & \cellcolor[rgb]{ .918,  .918,  .8} 0.44 (0.19) & 0.30 (0.40) & \cellcolor[rgb]{ .945,  .969,  .945} 0.34 (0.34) \\
    LOQ   & \cellcolor[rgb]{ .835,  .835,  .835} 0.56 (0.11) & \cellcolor[rgb]{ .867,  .867,  .686} 0.52 (0.14) & \cellcolor[rgb]{ .643,  .859,  .882} 0.54 (0.11) & \cellcolor[rgb]{ .722,  .843,  .722} 0.50 (0.15) \\
    LOX   & \cellcolor[rgb]{ .859,  .859,  .859} 0.52 (0.06) & \cellcolor[rgb]{ .89,  .894,  .741} 0.48 (0.07) & \cellcolor[rgb]{ .792,  .918,  .929} 0.44 (0.09) & \cellcolor[rgb]{ .82,  .898,  .82} 0.43 (0.09) \\
    \bottomrule
    \end{tabular}%
    }
\end{table}%

It is possible that hyperparameter tuning could recover these performance losses.
Anecdotally, we note that adding additional graph convolutional layers seems to help.
This can, however, dilute the explanations which become further ``de-localized'' with each convolution.

\section{Conclusion}

We describe and implement two new regularization techniques: Batch Representation Orthonormalization and Gini regularization.
We show qualitatively that these techniques disentangle node representations and force models to make predictions using fewer of them.

We demonstrate that models trained with these constraints generate better site attributions on a benchmark dataset for many attribution methods.
Further, we show that for models trained on assay data of interest to medicinal chemists, human experts significantly prefer attribution maps generated from models trained with the constraints introduced in this paper.

While we give an exact definition for the BRO constraint in Equation~\ref{eq:BRO}, the choice of the vector 2-norm is arbitrary.
The novelty of our approach is normalizing the node representations per graph, and it is likely that other norms may yield better results in other circumstances.
Similarly, while the Gini coefficient has desirable properties for this application, other measures of dispersion could also be explored.

Finally, our methods could be extended to graph architectures which exploit edge features.
Future work may explore the separate or joint orthonormalization of node and edge features.

\section*{Acknowledgements}

We would like to thank the Bayer medicinal chemists based in Berlin and Wuppertal for their essential feedback on our work.
We would also like to thank Andreas Goeller, Marco Bertolini, Jorge Kageyama, and Tuan Le for their helpful input.
Funding in direct support of this work: Bayer AG Life Science Collaboration (``Explainable AI'').

\bibliography{icml2021}
\bibliographystyle{icml2021}

\end{document}